\documentclass{article}

\usepackage{PRIMEarxiv}

\usepackage[utf8]{inputenc} % allow utf-8 input
\usepackage[T1]{fontenc}    % use 8-bit T1 fonts
\usepackage{hyperref}       % hyperlinks
\usepackage{url}            % simple URL typesetting
\usepackage{booktabs}       % professional-quality tables
\usepackage{amsmath,amssymb} % math environments and symbols (from original paper)
\usepackage{amsfonts}       % blackboard math symbols
\usepackage{nicefrac}       % compact symbols for 1/2, etc.
\usepackage{microtype}      % microtypography
\usepackage{fancyhdr}       % header
\usepackage{graphicx}       % graphics
\usepackage{multirow}       % multi-row table cells (from original paper)
\graphicspath{{media/}}     % organize your images and other figures under media/ folder

\title{Parallelism-Aligned Pruning and Round-Aware Context Reuse for On-Device Multimodal Agents
}

\author{
  Peipei Jiao, Mingkang Liu, Weiming Meng, Lichen Pang, Tianyuan Liu, and Junyan Li \\
  Juhaokan Technology Co., Ltd. \\
  Qingdao, China \\
  \texttt{cliff.pan@hisense.com} \\
}

\begin{document}
\maketitle

\begin{abstract}
On-device multimodal agents are limited by two deployment mismatches: pretrained attention structures may not divide evenly across the compute cores of an edge accelerator, causing irregular partitions; and standard prefix caching keeps historical visual tokens in the persistent KV cache, whose capacity grows with the number of rounds.
We present PARC, which combines parallelism-aligned structured pruning with lifetime-aware KV-cache reorganization.
At the model level, it uses first-order Taylor importance and grouped-query-attention-aware selection to make attention dimensions divisible by the target core count while preserving the query-to-key/value ratio and head dimension.
At runtime, it separates the system prompt, persistent text history, and transient visual observations by lifetime, updating text history during tool execution so that historical images leave the persistent cache and more interaction rounds fit a fixed budget.
On 2,062 GUI actions, the incremental interface shows no accuracy regression, and task-aware recovery training restores 98.7\% of the action-type accuracy gap and the coordinate hit rate to the original level.
Under seven-core deployment, hardware-aligned pruning reduces vision-encoder latency by 11.4\%, improves decoding throughput by 23.8\%, and lowers per-heap high-water memory by 12.7\%; a parameter-matched control confirms that the speedup stems mainly from core alignment rather than parameter reduction.
\end{abstract}

% keywords can be removed

\section{Introduction}

Vision--language models are evolving from static visual question answering into interactive agents that perceive an environment, generate structured actions, invoke tools, and respond to new observations, as in embodied navigation, mobile automation, and multi-round visual dialogue.
Compared with offline inference, such applications jointly require low time to first token (TTFT), high decoding throughput, accurate spatial grounding, and bounded device memory.

Two mismatches limit on-device efficiency.
First, general-purpose architectures are not necessarily aligned with the parallel resources of an edge accelerator: when a compiler partitions attention heads across compute cores, a non-divisible dimension creates uneven shards, and the slowest shard determines the synchronization latency.
Second, standard prefix caching treats the prefix as a contiguous sequence, so historical visual observations either keep consuming persistent cache capacity or, once removed, break direct reuse of the following text state; the KV-cache pressure thus grows with the number of rounds.

\begin{figure}[t]
    \centering
    \includegraphics[width=\columnwidth]{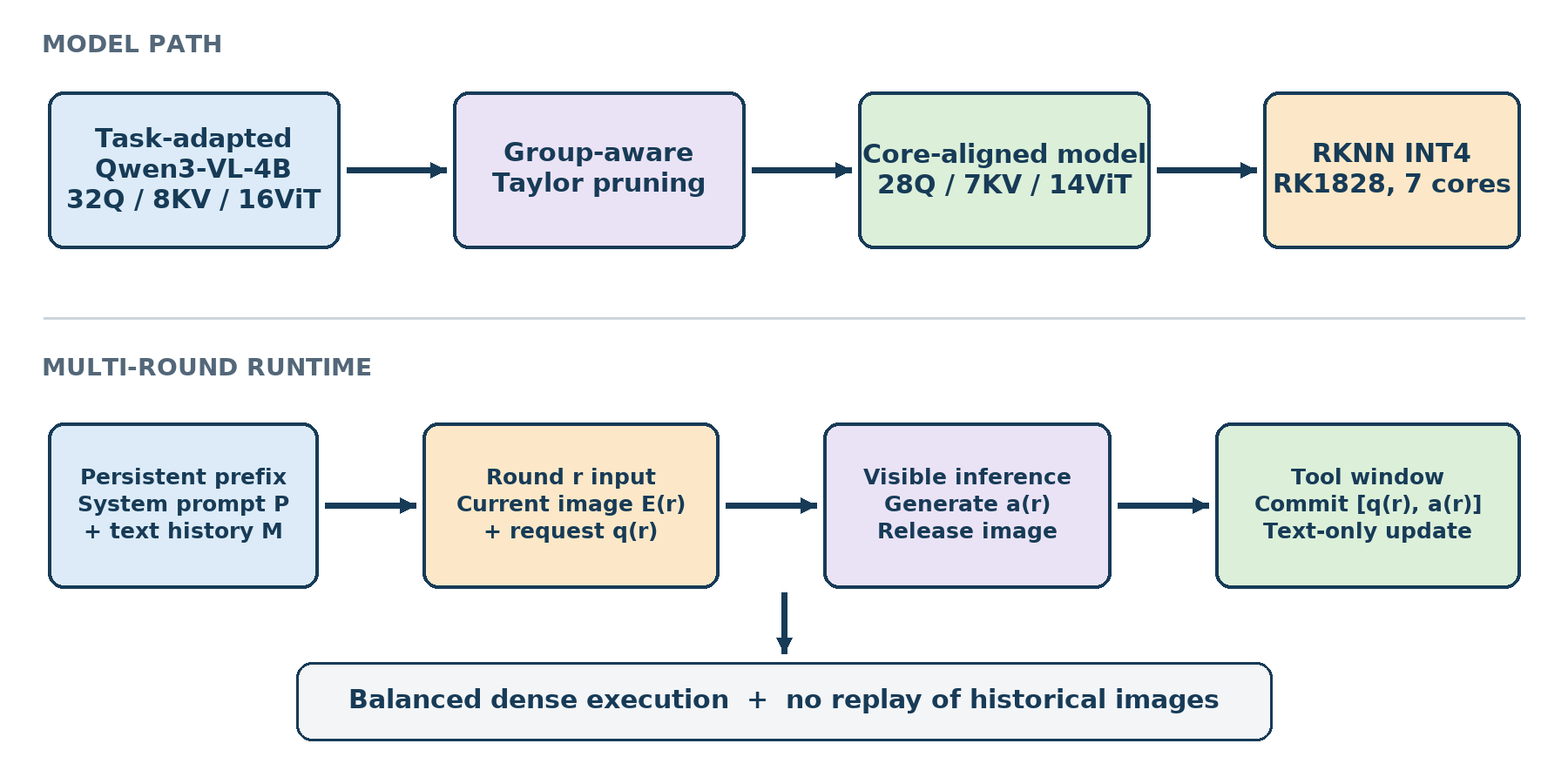}
    \caption{Overview of PARC. Model dimensions are aligned with available cores, while fixed prompts and round-specific inputs are managed according to their lifetimes.}
    \label{fig:overview}
\end{figure}

Existing pruning methods optimize parameter count or FLOPs and rarely treat the discrete core count as a structural constraint, while KV-cache research focuses on prefix reuse, eviction, or capacity rather than the differing lifetimes of prompts, text history, and images in tool-using agents.
We therefore propose PARC (Parallelism-Aligned Pruning and Round-Aware Context Reuse), illustrated in Fig.~\ref{fig:overview}.

The contributions of this work are threefold:
\begin{itemize}
    \item We formulate grouped-query-attention (GQA)-aware head pruning under a hardware parallelism constraint, preserving the GQA ratio, per-head dimension, and rotary-position compatibility while producing more regular partition candidates and reducing the risk of uneven shards or local serialization.
    \item We introduce a round-aware inference schedule that separates persistent text from transient images and commits new text history during tool execution, thereby reorganizing KV-cache state by context lifetime, reducing persistent visual-token occupancy, and extending the feasible interaction length under a fixed cache budget.
    \item We conduct deployment experiments on a seven-core edge NPU and systematically evaluate direct pruning, recovery training, and multi-round incremental inference on 2,062 GUI actions, using accuracy and device measurements to quantify effects on visual encoding, context prefill, autoregressive decoding, and memory usage.
\end{itemize}

\section{Related Work}

\subsection{Structured Pruning and Hardware-Aware Compression}

Unstructured methods such as SparseGPT remove individual weights and can achieve high sparsity \cite{frantar2023sparsegpt}.
Their realized speedup, however, depends on sparse storage formats and specialized kernels.
Structured pruning removes complete channels, layers, heads, or tensor groups and retains dense regular tensors that are more readily supported by existing conversion and operator-fusion pipelines.

The Taylor criterion estimates the loss change caused by removing a structure using parameter--gradient products.
DepGraph and Torch-Pruning explicitly track operator dependencies so that coupled input and output dimensions are pruned consistently \cite{fang2023depgraph}.
Prior studies commonly target FLOPs, parameter count, or a uniform pruning ratio.
PARC instead introduces the accelerator core count into the feasible set of attention dimensions and optimizes partition regularity in addition to model size.

\subsection{KV-Cache Optimization}

Autoregressive transformers create key and value tensors during prefill and repeatedly read them during decoding \cite{vaswani2017attention}.
H2O retains recent and high-importance tokens \cite{zhang2023h2o}, while SnapKV selects salient cache positions from an observation window near the end of a prompt \cite{li2024snapkv}.
PagedAttention improves memory utilization through non-contiguous logical pages \cite{kwon2023pagedattention}, and Gist Tokens compress long prompts into reusable representations \cite{mu2023gist}.
These methods mainly decide which cache entries to retain or how to store them.
PARC is complementary: it changes prefill boundaries and separates fixed, accumulating, and transient context segments.

\subsection{Interactive Multimodal Agents}

ReAct interleaves language reasoning with external actions \cite{yao2023react}.
Platforms such as LEGENT provide continuous perception, language interaction, action execution, and trajectory generation \cite{cheng2024legent}.
Most related systems focus on agent capability and environment construction, with inference performed on server GPUs.
On-device deployment introduces a tighter coupling among visual encoding, text prefill, cache storage, and tool scheduling.
PARC addresses this systems-level coupling without assuming a specific graphical interface.

\section{Problem Formulation}

\subsection{Model--Hardware Parallelism Mismatch}

Consider a multimodal transformer deployed on an accelerator with $C$ available cores.
The model exposes several potentially parallel dimensions, including language query heads $H_q$, language key/value heads $H_{kv}$, and vision heads $H_v$.
The compiler may partition one or more of these dimensions across the cores.
When a partition dimension is not divisible by $C$, the resulting workload can contain an uneven final shard.
For example, assigning eight equal-cost groups to seven cores requires one core to execute two groups or requires the compiler to introduce padding.
Because synchronized operators complete at the pace of the slowest core, reducing the average work is insufficient if the critical shard remains unchanged.

The structural objective is therefore not merely to minimize the number of heads.
Given an admissible accuracy loss $\epsilon$, we seek target dimensions
\begin{equation}
\mathcal{H}^{\star}=
\left(H_q^\star,H_{kv}^\star,H_v^\star\right)
\label{eq:targettuple}
\end{equation}
that minimize measured device latency while satisfying divisibility, GQA consistency, and accuracy constraints:
\begin{align}
H_q^\star \bmod C &= 0, &
H_{kv}^\star \bmod C &= 0, \nonumber\\
H_v^\star \bmod C &= 0, &
\frac{H_q^\star}{H_{kv}^\star} &=
\frac{H_q}{H_{kv}}, \label{eq:constraints}\\
\Delta\mathcal{A}(\mathcal{H}^{\star}) &\leq \epsilon. &&
\nonumber
\end{align}
Here, $\mathcal{A}$ denotes task accuracy.
The final constraint cannot be established from structure alone and motivates recovery fine-tuning and task-level evaluation.

\subsection{Redundant Multi-Round Prefill}

Let $L_P$, $L_E^{(r)}$, $L_q^{(r)}$, and $L_a^{(r)}$ be the token lengths of the system prompt, visual observation, request, and response at round $r$.
If every round serializes the complete multimodal history without separating transient images from persistent text, the number of tokens presented to prefill over $R$ rounds is approximately
\begin{equation}
N_{\mathrm{full}}=
\sum_{r=1}^{R}
\left[
L_P+
\sum_{i=1}^{r}L_E^{(i)}+
\sum_{i=1}^{r}L_q^{(i)}+
\sum_{i=1}^{r-1}L_a^{(i)}
\right].
\label{eq:fullprefill}
\end{equation}
The repeated-image term grows quadratically with $R$ when the per-round observation length is approximately constant.
This growth is particularly expensive because a visual observation must first pass through the vision encoder and is then represented by many visual tokens in the language model.

With successful prefix reuse and transient visual observations, the number of newly prefilled tokens is approximated by
\begin{equation}
N_{\mathrm{PARC}}\approx
L_P+\sum_{r=1}^{R}
\left(L_E^{(r)}+L_q^{(r)}+L_a^{(r)}\right).
\label{eq:parcprefill}
\end{equation}
This counts newly computed prefill tokens; previously cached text still participates in attention, so cache reads and KV capacity still grow with conversation length.

\subsection{KV-Cache Capacity Under a Fixed Budget}

Let $B$ denote the KV-cache capacity in tokens, with approximately constant per-round lengths $L_E$, $L_q$, $L_a$.
Retaining all historical images bounds the number of complete rounds by $R_{\mathrm{full}}=\lfloor(B-L_P)/(L_E+L_q+L_a)\rfloor$, whereas keeping only the current image resident gives $R_{\mathrm{PARC}}=\lfloor(B-L_P-L_E)/(L_q+L_a)\rfloor$.
For example, with $B=3072$, $L_P=1024$, $L_E=220$, and $L_a=100$, full retention supports six rounds and transient retention up to 18.
This is an analytical capacity illustration, not a measured device limit, and is distinct from the measured device-memory footprint reported in Table~\ref{tab:memory}.

\subsection{Joint Objective}

PARC targets foreground interaction latency, decomposed per round into vision, prefill, decode, and scheduling terms.
Structural pruning primarily affects the first three, whereas round-aware reuse reduces redundant prefill and overlaps history maintenance with tool execution; validating the method therefore requires jointly measuring accuracy, per-stage latency, decoding throughput, and cache usage.

\section{Method}

\subsection{Parallelism Constraint}

\begin{figure}[t]
    \centering
    \includegraphics[width=\columnwidth]{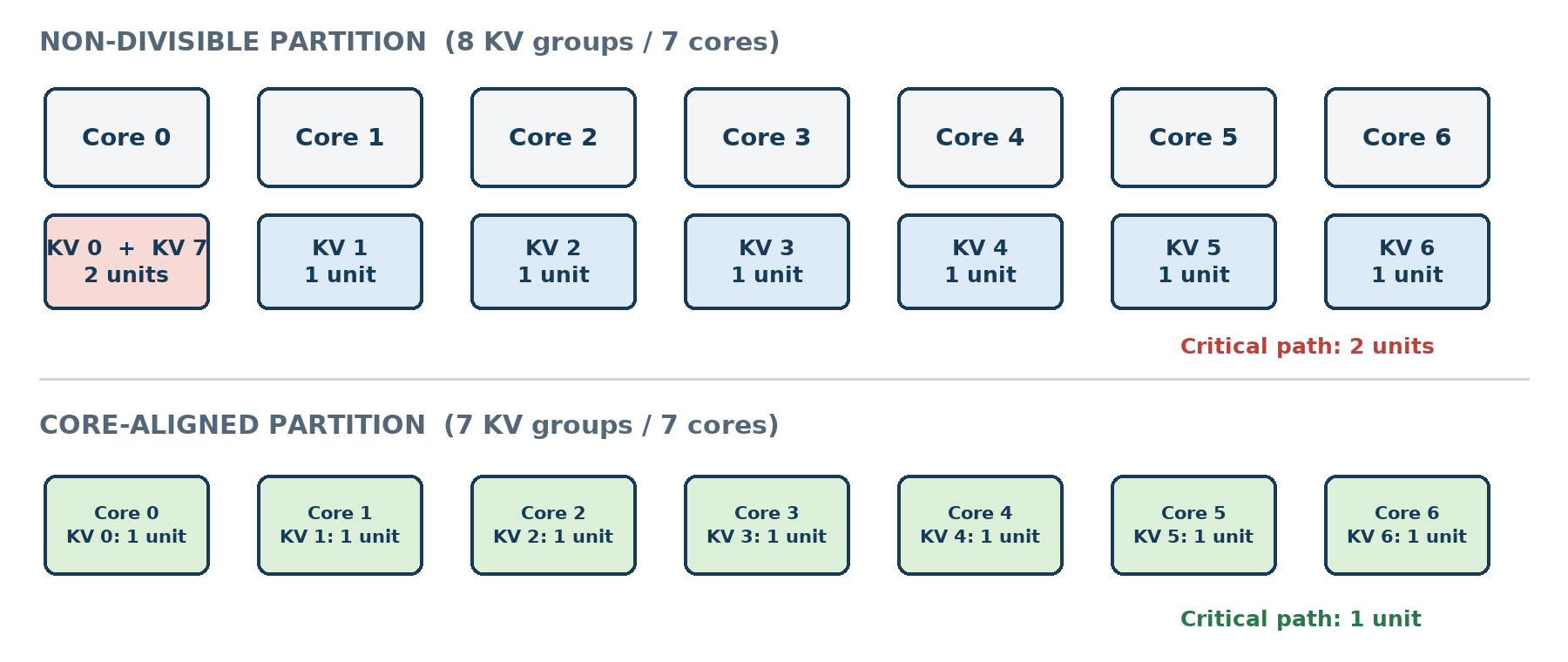}
    \caption{Illustrative non-divisible and aligned mappings. The method generalizes to other core counts and partition granularities.}
    \label{fig:mapping}
\end{figure}

Let $H$ be the number of approximately equal-cost attention partitions and $C$ the number of available accelerator cores.
Under static uniform partitioning, the slowest core processes $\lceil H/C\rceil$ partitions.
We define the discrete partition utilization as
\begin{equation}
\eta(H,C)=\frac{H}{C\lceil H/C\rceil}.
\label{eq:utilization}
\end{equation}
If $H$ is not divisible by $C$, some cores receive an additional partition, so a target $H^\star$ with $H^\star\bmod C=0$ is a candidate for regular mapping (Fig.~\ref{fig:mapping}).
Equation~\eqref{eq:utilization} is a structural screening metric, not a predicted end-to-end speedup; the realized mapping must still be verified from the compiled graph.

\subsection{GQA-Aware Language-Head Pruning}

Let $H_q$ and $H_{kv}$ denote the original query- and key/value-head counts, with GQA ratio $g=H_q/H_{kv}$.
For $C$ available cores, PARC selects
\begin{equation}
H_{kv}^{\star}=mC,\qquad H_q^{\star}=gH_{kv}^{\star},
\label{eq:targets}
\end{equation}
where $m$ is a positive integer selected under the accuracy or latency budget.
This construction makes both target dimensions divisible by $C$ while preserving $g$.

For query head $h$ in layer $l$, we use the first-order Taylor score
\begin{equation}
I_{l,h}=\operatorname{mean}\left|
W^{Q}_{l,h}\odot
\frac{\partial\mathcal{L}}{\partial W^{Q}_{l,h}}
\right|.
\label{eq:taylor}
\end{equation}
Scores are aggregated over the $g$ query heads associated with each KV head.
The highest-scoring KV groups are retained subject to \eqref{eq:targets}.
The output dimensions of $W^Q$, $W^K$, and $W^V$ and the corresponding input dimension of $W^O$ are sliced consistently.
When gradients are unavailable, the score falls back to the $\ell_2$ norm.
Layer count, residual width, vocabulary, and per-head rotary dimension remain unchanged.

\subsection{Vision-Head Pruning}

For an original vision-head count $H_v$, the target is $H_v^\star=nC$, where $n$ is a positive integer and $H_v^\star\leq H_v$.
Heads are ranked using the magnitude of their query projection.
The corresponding rows of the fused QKV projection and columns of the output projection are removed together.
The per-head dimension and cross-modal output width remain unchanged, limiting changes to downstream interfaces.

\subsection{Dependency-Consistent Pruning Procedure}

Head removal changes multiple coupled tensor dimensions.
For each language layer, PARC aggregates query-head importance within the original GQA groups and retains the highest-scoring KV groups with their query heads; with per-head dimension $d_h$, the retained indices slice $W^Q$/$W^O$ to $H_q^\star d_h$ and $W^K$/$W^V$ to $H_{kv}^\star d_h$, together with any bias vectors.
For a vision block with a fused projection $W^{QKV}\in\mathbb{R}^{3H_vd_v\times d_{\mathrm{model}}}$, the same selected-head indices are applied to the query, key, and value ranges and to the matching output-projection columns.
This preserves the block output width, leaving the residual path and cross-modal connector unchanged.
After pruning, PARC updates the model metadata and validates the result in three stages: projection shapes and divisibility, a finite floating-point forward pass, and successful target-backend conversion of all modified operators.
These checks distinguish a mathematically valid head selection from a deployable model artifact.

\subsection{Backend Conversion and Deployment Adaptation}

The pruned floating-point checkpoint is first validated in the source framework and is then converted and quantized for the target backend.
The key requirement is backend independent: because the query, KV, and vision projection widths differ from the original architecture, any target runtime must read the updated configuration and preserve the pruned tensor dimensions and the fixed per-head encoding dimensions, rather than reconstructing them from the original model shape (for example, by inferring the per-head dimension from hidden width divided by head count).
Platform-specific conversion details for the RKNN/Qwen3-VL path used in this work are given in Section~\ref{sec:results}.

\subsection{Target Selection and Recovery}

The feasible targets in \eqref{eq:constraints} form a discrete set.
For a fixed core count, PARC enumerates candidates that preserve the original GQA ratio and per-head dimension.
Each candidate can be screened using parameter count and the utilization proxy in \eqref{eq:utilization}, but the final choice should be based on measured latency and task accuracy.
An important control is a model with a similar parameter count but a non-divisible head configuration; it separates the benefit of removing parameters from the benefit of producing regular partitions.

Removing attention heads changes the learned subspace even when tensor interfaces remain compatible.
We therefore treat recovery fine-tuning as part of the complete deployment pipeline rather than an optional post-processing step.
Either low-rank adaptation \cite{hu2022lora} or full-parameter training can be used on structured action, spatial-grounding, and tool-use examples.
The recovery set should contain coordinate-sensitive examples because the results in Section~\ref{sec:results} show that localization is substantially more fragile than coarse action recognition.

\subsection{Round-Aware Incremental Context Reuse}

Let $P$ denote the system prompt, and let $E^{(r)}$, $q^{(r)}$, and $a^{(r)}$ denote the visual observation, user request, and model response at round $r$.
The persistent textual history before round $r$ is
\begin{equation}
M^{(r-1)}=[q^{(1)},a^{(1)},\ldots,q^{(r-1)},a^{(r-1)}].
\label{eq:history}
\end{equation}
$P$ is fixed for the task, $M^{(r-1)}$ grows across rounds, and $E^{(r)}$ is transient.
Past visual observations are not stored in persistent history.

At initialization, the runtime prefills $P$ once and stores $KV(P)$.
At visible round $r$, the business layer submits $[E^{(r)},q^{(r)}]$ in \emph{history-retention mode}.
The runtime reuses the retained text-only state $KV(P,M^{(r-1)})$, appends the current visual and query tokens, and generates $a^{(r)}$.
The cache produced by this call is transient because it contains $E^{(r)}$.

\begin{figure}[t]
    \centering
    \includegraphics[width=\columnwidth]{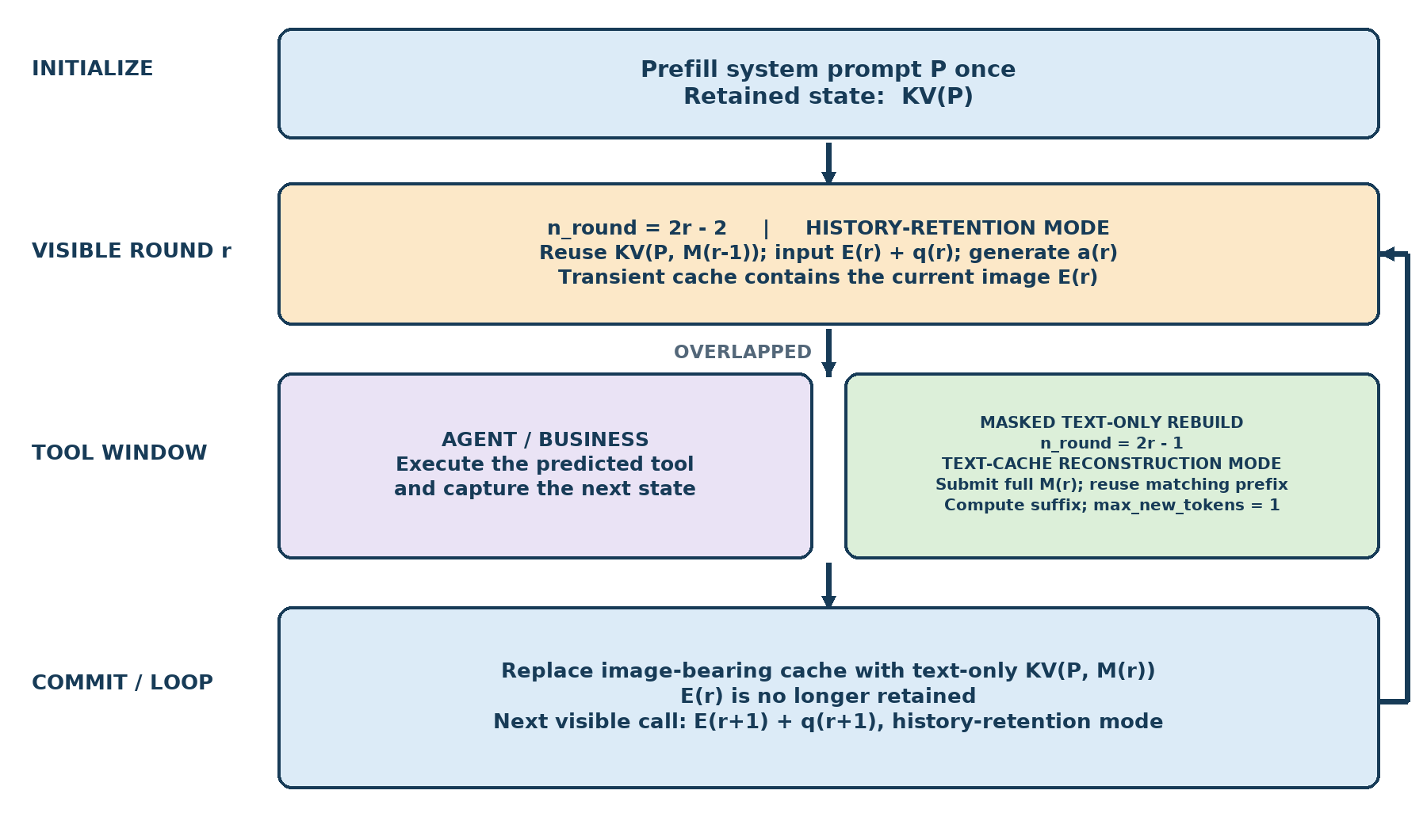}
    \caption{Alternating visible inference and masked text-only KV-cache reconstruction.}
    \label{fig:timeline}
\end{figure}

After $a^{(r)}$ is produced, the business layer starts the requested tool and serializes the complete cumulative text history
$M^{(r)}=[q^{(1)},a^{(1)},\ldots,q^{(r)},a^{(r)}]$.
During the tool-execution window, it submits $M^{(r)}$ in a text-only maintenance call in \emph{text-cache reconstruction mode} with minimal generation.
In this mode, the previous cache is not retained as the final session state; nevertheless, the runtime performs token-prefix matching and directly reuses the matching prefix KV entries, computing only the unmatched suffix.
The call therefore replaces the transient image-bearing state with $KV(P,M^{(r)})$ and removes $E^{(r)}$ from the persistent cache.
Its output is masked and is not exposed as another assistant response.
The next visible call returns to history-retention mode with $[E^{(r+1)},q^{(r+1)}]$, and the procedure repeats.
Figure~\ref{fig:timeline} summarizes the state transition.

Through prefix matching, this schedule avoids recomputing the system prompt, previous images, and committed text, and prevents historical visual tokens from accumulating in the persistent cache, freeing budget for more interaction rounds.
It does not eliminate textual-history growth, so the realized benefit depends on prefix-hit rate, visual-token count, and tool duration.

\subsection{Runtime Workflow and State Management}

The prototype has four logical layers: an application layer (screen capture, request formatting, action execution, task boundaries), a round-aware scheduler that owns the textual history and chooses between visible inference and masked maintenance, the multimodal runtime exposing session-level KV state, and the target backend on the NPU.
As shown in Fig.~\ref{fig:timeline}, a visible round submits the current image and request and releases the image buffer after vision inference, while a masked maintenance call during tool execution reconstructs the text-only persistent state; the new logical round is committed only after it succeeds.
Two details are essential for correctness: the appended history must use exactly the same tokenizer and chat delimiters as visible inference, or prefix matching fails; and the authoritative history stays in the application layer, so the runtime cache can be rebuilt after any error or eviction.

\subsection{Cache-State Invariants and Failure Handling}

Immediately after visible inference, the transient cache may contain $E^{(r)}$.
After the masked maintenance call, the logical persistent state must represent
\begin{equation}
\mathcal{K}^{(r)}=KV\!\left(P,M^{(r)}\right),
\label{eq:cacheinvariant}
\end{equation}
while no previous observation $E^{(i)}$, $i\leq r$, is part of the persistent logical input.
Equation~\eqref{eq:cacheinvariant} is a post-maintenance state invariant: a new visual observation may be processed transiently, but the scheduler returns to a text-only persistent state before the next round.
This distinction prevents image history from growing without removing textual evidence required for subsequent decisions.

The maintenance call is useful only when it overlaps tool execution: if the tool finishes first, the next observation waits for the update or a synchronized fallback; if prefix matching fails or inference is interrupted, the state is rebuilt from the authoritative history and the logical round is not advanced until the update succeeds.
At task completion, cancellation, or eviction, the business layer clears both the runtime KV state and its history metadata, avoiding silent divergence between the visible conversation and the internal cache.

\section{Experiments and Results}\label{sec:results}

\subsection{Experimental Setup}

We use a task-adapted Qwen3-VL-4B checkpoint \cite{qwen3vl}, obtained by merging a balanced mixture of GUI-task LoRA adapters, and deploy it to an RK1828 edge NPU.
The original language module contains 32 query heads and eight KV heads, and the vision encoder contains 16 attention heads.
PARC changes these counts to 28, seven, and 14, respectively.
Taylor importance is estimated on 128 WikiText-2 calibration sequences of length 2,048.
The resulting checkpoint retains all 36 language layers and contains 4.31 billion parameters, compared with 4.4378 billion before pruning, a reduction of approximately 128 million parameters (2.9\%).

The checkpoint is converted and quantized using the RKNN toolchain under a W4A16 configuration (4-bit weights and 16-bit activations).
Unless stated otherwise, the ``Pruned'' model in the device measurements of Tables~\ref{tab:deviceperf} and~\ref{tab:memory} refers to the deployed pruned-and-recovered checkpoint after this conversion and quantization, not the directly pruned model without recovery.
For this Qwen3-VL implementation, the vision QKV and output projections use the explicit retained width $H_v^\star d_v$ rather than the original hidden width.
The vision RoPE path reads the fixed 64-dimensional head size from the updated configuration, avoiding the invalid inference $1024/14$ after pruning, while the rotary dimension remains 32; the text RoPE path also handles a missing \texttt{rope\_scaling} field.
The runtime's history-control flag maps history-retention and text-cache reconstruction modes to values 1 and 0, respectively.
Before task evaluation, identical text-only and image--text inputs are used to check the source and converted models for functional consistency.

In this work, recovery is performed with LoRA applied to all linear projections, using rank 8, scaling factor 16, and dropout 0.05.
We train for two epochs with a cosine schedule, a peak learning rate of $1\times10^{-5}$, 50 warmup steps, an effective batch size of 128, a maximum sequence length of 8{,}192, and bf16 precision.
The recovery data is disjoint from the 2,062 evaluation actions.
In our runs, the procedure is not sensitive to the recovery-set size: training on 5{,}000 and on 50{,}000 examples yields comparable recovered accuracy.

We compare the original, directly pruned, and recovered models using both one-shot inference and the multi-round GUI incremental interface.
All accuracy runs use the same generation configuration.

The test set contains 2,062 GUI-action steps evaluated under the multi-round GUI workflow and covers click, application launch, text entry, wait, back, and swipe actions.
We report action-type accuracy, coordinate hit rate, and exact-match accuracy.
Coordinate hit rate evaluates whether a predicted position falls inside the target region or satisfies a distance threshold.
The configured distance threshold is 50.0.
It is computed over samples with a valid coordinate prediction, whose count is reported in parentheses.
For each model checkpoint, the same action set and generation configuration are evaluated using both conventional one-shot inference and the GUI incremental workflow.
The paired aggregate metrics therefore measure whether multi-round state management changes task accuracy.

\subsection{Multi-Round GUI Accuracy}

\begin{table}[t]
\caption{Accuracy on 2,062 GUI Actions}
\label{tab:accuracy}
\centering
\resizebox{\columnwidth}{!}{
\begin{tabular}{lclccc}
\toprule
Model & NPU Cores & Interface & Type Acc. & Coord. Hit & Exact Match\\
\midrule
Original & 8 & Baseline & 83.4 & 85.2 (757) & 77.9\\
Original & 7 & One-shot & 83.1 & 85.7 (759) & 77.3\\
Original & 7 & Multi-round GUI & 83.1 & 85.7 (759) & \textbf{77.4}\\
Pruned & 7 & One-shot & 67.1 & 36.9 (759) & 43.8\\
Pruned & 7 & Multi-round GUI & 67.4 & 36.4 (758) & 43.8\\
Recovered & 7 & One-shot & \textbf{83.0} & \textbf{86.3 (759)} & 71.5\\
Recovered & 7 & Multi-round GUI & 82.9 & 85.8 (758) & 71.2\\
\bottomrule
\end{tabular}}
\end{table}

As shown in Table~\ref{tab:accuracy}, all aggregate metrics differ by at most 0.5 percentage points between one-shot and multi-round GUI inference across the three model checkpoints, indicating no material accuracy regression from the multi-round interface.
The eight-core baseline in Table~\ref{tab:accuracy} is close to the original model's seven-core results, so it is reported as a deployment reference rather than evidence that core count changes model accuracy.

\begin{figure}[t]
    \centering
    \includegraphics[width=\columnwidth]{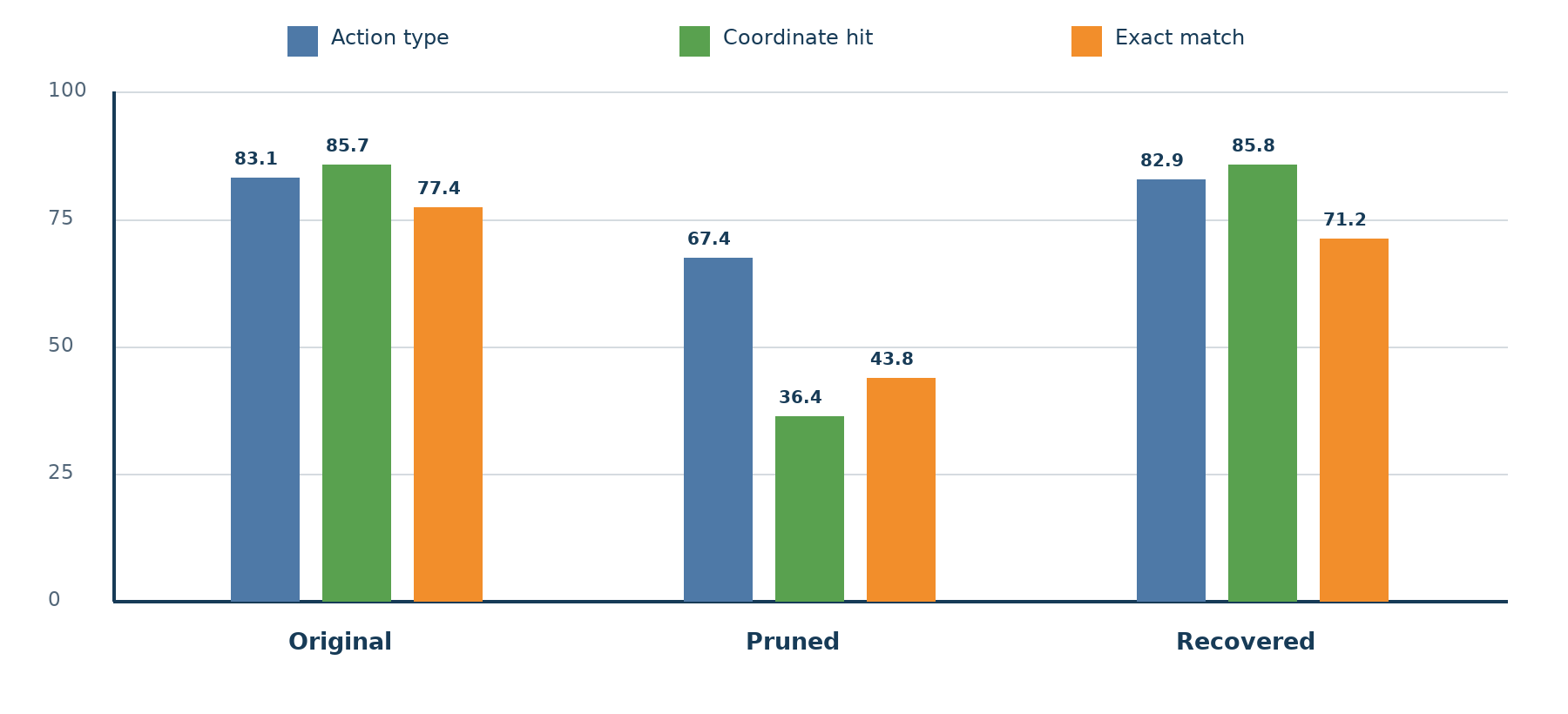}
    \caption{Incremental-interface accuracy before pruning, after direct pruning, and after recovery training. Coordinate hit rate is conditional on a valid coordinate prediction.}
    \label{fig:accuracybars}
\end{figure}

Direct pruning causes a substantial loss.
Under the incremental interface, action-type accuracy, coordinate hit rate, and exact match decrease by 15.7, 49.3, and 33.6 percentage points, respectively.
Recovery training raises these metrics to 82.9\%, 85.8\%, and 71.2\%.
Relative to the original incremental model, it closes 98.7\% of the action-type accuracy gap and 81.5\% of the exact-match gap.
Under the incremental interface, the recovered coordinate hit rate reaches 85.8\%, essentially matching the original model's 85.7\%, so coordinate localization is restored to the original-model level.
The residual gap relative to the original model is concentrated in exact match (71.2\% versus 77.4\%) rather than in action-type accuracy or coordinate hit rate.
These results show that hardware-aligned structure alone does not preserve model ability: direct pruning affects spatial grounding more strongly than aggregate action recognition, but recovery training closes both the action-type and coordinate gaps.
Structural regularity and task accuracy must therefore be treated as separate objectives, with task-relevant recovery training included in the deployment pipeline.

\subsection{Model Inference Performance}

We report system-prompt loading time, visual encoding latency, language-model prefill throughput, and autoregressive decoding throughput.
For $N_{\mathrm{out}}$ generated tokens and decoding duration $T_{\mathrm{decode}}$,
\begin{equation}
\mathrm{TPS}=\frac{N_{\mathrm{out}}}{T_{\mathrm{decode}}},
\qquad
\mathrm{TPOT}=\frac{T_{\mathrm{decode}}}{N_{\mathrm{out}}}.
\label{eq:decode}
\end{equation}
Table~\ref{tab:deviceperf} reports RKNN inference measurements for the original model using eight and seven NPU cores, for the aligned pruned model (28/7/14 heads) using seven cores, and for a parameter-matched non-aligned control described below.
Each reported value is the mean of ten repeated runs under the same configuration.

\begin{table}[t]
\caption{RKNN Model Inference Performance}
\label{tab:deviceperf}
\centering
\resizebox{\columnwidth}{!}{
\begin{tabular}{llcrrrrr}
\toprule
Model & NPU & Params & System prompt & Vision & ViT & Prefill & Decode\\
 & cores & & (s) & input & (ms) & (tok/s) & (tok/s)\\
\midrule
Original & 8 & 4.4378B & 0.848 & $640\times352$ & 184.84 & 729.59 & 79.17\\
Original & 7 & 4.4378B & 1.24 & $640\times352$ & 223.67 & 650.59 & 62.32\\
Aligned pruned & 7 & 4.31B & 1.04 & $640\times352$ & 198.28 & 651.04 & \textbf{77.12}\\
FFN-pruned (non-aligned) & 7 & 4.3140B & 1.23 & $640\times352$ & 225.72 & 655.00 & 62.89\\
\bottomrule
\end{tabular}}
\end{table}

Relative to the original seven-core configuration, the aligned pruned seven-core model reduces ViT latency from 223.67 to 198.28~ms (11.4\%) and increases decoding throughput from 62.32 to 77.12~tokens/s (23.8\%).
Prefill throughput is essentially unchanged (650.59 to 651.04~tokens/s); prefill is compute-bound and is therefore far less sensitive to the head-count reduction than decoding, which explains why it does not improve.

To separate the effect of parameter reduction from the effect of core alignment, we add a parameter-matched non-aligned control.
Starting from the original model, this control keeps the attention head configuration unchanged (32 query and eight key/value heads, which are not divisible by seven), and instead reduces parameters only in the feed-forward intermediate dimension using the same first-order Taylor criterion, reaching 4.3140~billion parameters, essentially the same size as the aligned pruned model (4.31~billion).
Despite this matched parameter budget, the non-aligned control attains only 62.89~tokens/s of decoding throughput and 225.72~ms of ViT latency, close to the original seven-core model and far from the aligned pruned model's 77.12~tokens/s and 198.28~ms.
Because the two pruned models share essentially the same parameter count and differ mainly in whether the head count is divisible by the core count, the roughly 22.6\% decoding-throughput gap between them is attributable to core alignment rather than to parameter reduction.
This supports the central claim that aligning attention dimensions with the core count, not merely reducing head count, drives the measured speedup.
Divisibility remains a structural precondition for regular partitioning.
The current RKNN toolchain does not expose per-core execution traces, so we rely on the parameter-matched control above together with the per-heap memory distribution in Section~\ref{sec:results} as indirect evidence of the mapping, and leave direct per-core tracing to future work.

\subsection{Memory Usage}

Before model loading, the \texttt{max used size} of every NPU heap is 96 bytes, providing a cleared baseline for both measurements.
Table~\ref{tab:memory} reports the per-heap high-water values observed during inference.

\begin{table}[t]
\caption{NPU Heap High-Water Usage During Inference}
\label{tab:memory}
\centering
\begin{tabular}{lrrr}
\toprule
Heap & Original (MB) & Pruned (MB) & Reduction (\%)\\
\midrule
n0 & 524.93 & 470.39 & 10.4\\
n1 & 476.71 & 416.72 & 12.6\\
n2 & 480.98 & 415.42 & 13.6\\
n3 & 481.64 & 410.54 & 14.8\\
n4 & 474.45 & 410.56 & 13.5\\
n5 & 468.04 & 409.93 & 12.4\\
n6 & 462.80 & 409.28 & 11.6\\
n7 & 0.18 & 0.18 & 0.0\\
\midrule
Total & 3369.73 & 2943.03 & 12.7\\
\bottomrule
\end{tabular}
\end{table}

The aggregate of the per-heap high-water values decreases by 426.70~MB, from 3.370 to 2.943~GB, corresponding to a 12.7\% reduction under the reported RKNN memory metric.
All compute heaps n0--n6 show reductions of 10.4--14.8\%, while the small n7 management heap remains unchanged.
The inference load is carried by the seven compute heaps n0--n6 with comparable footprints, while the management heap n7 stays negligible, which is consistent with execution partitioned across the seven cores; this memory view is a coarse indicator only, and direct per-core execution traces remain the definitive way to confirm the mapping.
Because maxima from different heaps are not guaranteed to occur at exactly the same instant, the aggregate should be interpreted as a sum of per-heap high-water marks rather than a synchronized process-level peak.

\subsection{Round-Aware Runtime Behavior}

To examine how the round-aware schedule behaves across rounds, we run the aligned pruned model on seven cores under two strategies and track both the feasible number of rounds and the per-round cost.
In the full-history strategy, each visible round prepends the accumulated history, including all previous images, to the current image and query.
In the round-aware strategy, each visible round processes only the current image and query, while the growing textual history is committed by a masked maintenance call issued during the tool-execution window; the maintenance call is labeled ``round-aware (mask)'' and produces no visible response.
Table~\ref{tab:roundaware} reports, for the first six rounds, the per-round visual input, prefill time (used here as the time to first token, TTFT), and prefill throughput.

\begin{table}[t]
\caption{Per-Round Runtime Behavior on Seven Cores (Aligned Pruned Model)}
\label{tab:roundaware}
\centering
\resizebox{\columnwidth}{!}{
\begin{tabular}{cllrrr}
\toprule
Round & Strategy & Vision & Input tok. & TTFT (ms) & Prefill (tok/s)\\
\midrule
\multirow{3}{*}{1} & full & $640\times352$ & 238 & 302.09 & 787.85\\
 & round-aware & $640\times352$ & 238 & 302.48 & 786.84\\
 & round-aware (mask) & -- & 627 & 140.48 & 4463.43\\
\midrule
\multirow{3}{*}{2} & full & $640\times352$ & 236 & 321.30 & 734.51\\
 & round-aware & $640\times352$ & 236 & 310.69 & 759.59\\
 & round-aware (mask) & -- & 644 & 144.23 & 4465.18\\
\midrule
\multirow{3}{*}{3} & full & $640\times352$ & 240 & 332.09 & 722.70\\
 & round-aware & $640\times352$ & 240 & 314.18 & 763.90\\
 & round-aware (mask) & -- & 665 & 146.30 & 4545.36\\
\midrule
\multirow{3}{*}{4} & full & $640\times352$ & 235 & 338.95 & 693.32\\
 & round-aware & $640\times352$ & 235 & 314.38 & 747.50\\
 & round-aware (mask) & -- & 681 & 149.25 & 4562.75\\
\midrule
\multirow{3}{*}{5} & full & $640\times352$ & 241 & 353.36 & 682.02\\
 & round-aware & $640\times352$ & 241 & 315.91 & 762.87\\
 & round-aware (mask) & -- & 703 & 151.97 & 4625.97\\
\midrule
\multirow{3}{*}{6} & full & $640\times352$ & 236 & 359.70 & 656.10\\
 & round-aware & $640\times352$ & 236 & 315.55 & 747.89\\
 & round-aware (mask) & -- & 720 & 154.89 & 4648.52\\
\bottomrule
\end{tabular}}
\end{table}

The primary benefit of the schedule is capacity, not per-round speed.
Under the full-history strategy, every visible round retains all previous images in the persistent KV cache, which therefore fills quickly: on this device the cache budget is exhausted at the sixth round, so roughly six rounds are the practical limit.
Under the round-aware strategy, historical visual tokens are removed from the persistent cache and only the current image is resident, so the same fixed cache budget sustains about twenty interaction rounds, more than a $3\times$ increase, without changing the per-round inference procedure.
This measured six-versus-twenty-round gap confirms the token-budget analysis in Section~III-C on the actual device.
Crucially, this capacity gain does not cost extra per-round latency.
The visible-round TTFT of the round-aware strategy stays essentially flat (302.48 to 315.55~ms) and matches that of full history at the rounds where full history is still feasible, so users see no slowdown.
The growing textual history is committed by the masked maintenance call issued during tool execution; its input grows across rounds while its prefix is reused (its prefill throughput exceeds 4{,}400~tokens/s, indicating that only the unmatched suffix is recomputed), and because it runs inside the tool-execution window its cost is hidden from the user-facing path rather than added to it.
In short, the schedule extends the feasible interaction length under a fixed cache budget while keeping foreground latency unchanged.

\subsection{Discussion and Limitations}

PARC combines two optimizations that act at different stages: parallelism-aligned pruning changes model execution, whereas lifetime-aware KV-cache reorganization reduces persistent visual-token retention and repeated multimodal prefill.
Explicit KV-cache occupancy and prefix-hit-rate readouts, not exported by the current toolchain, would further quantify the round-aware schedule, and we leave these measurements to future work.

The present study covers one 4B-scale vision--language model, one edge-NPU platform, and GUI actions.
Device latency and throughput are reported as the mean of ten repeated runs per configuration; the parameter-matched non-aligned control in Section~\ref{sec:results} already isolates core alignment from parameter reduction as the main source of the measured speedup, and direct per-core execution traces, which the current toolchain does not expose, would provide further confirmation of the physical mapping.
The dataset is also imbalanced across action types, so aggregate accuracy can conceal minority-class behavior; class-wise results or balanced metrics would provide a more complete assessment.

Finally, round-aware reuse does not eliminate textual-history growth, and it assumes that request--response text is sufficient after each visual round.
Tasks that require details visible only in an earlier image may need selective visual-memory retention or a textual visual summary, introducing an additional accuracy--memory trade-off.

\section{Conclusion}

We introduced PARC, a framework that jointly addresses irregular core partitioning and KV-cache lifetime mismatch through parallelism-aligned structured pruning and round-aware context reuse.
By retaining only textual history between visual rounds, PARC also reduces persistent KV-cache occupancy and increases the feasible number of interactions under a fixed cache budget.
On 2,062 GUI actions evaluated in the multi-round workflow, the incremental interface produces accuracy comparable to one-shot inference, while direct pruning without recovery training substantially degrades fine-grained localization.
Recovery training then closes 98.7\% and 81.5\% of the action-type and exact-match gaps, restoring coordinate localization to the original-model level.

In the seven-core edge-NPU comparison, the aligned pruned model reduces ViT latency from 223.67 to 198.28~ms and increases decoding throughput from 62.32 to 77.12~tokens/s, while prefill throughput remains nearly unchanged.
Aggregate per-heap high-water memory usage decreases from 3.370 to 2.943~GB (12.7\%).
A parameter-matched non-aligned control, which prunes only the feed-forward dimension and keeps the original attention heads, reaches the same parameter budget (4.314 versus 4.31~billion) yet attains only 62.89~tokens/s, close to the unpruned model; this isolates core alignment, rather than parameter reduction alone, as the main source of the measured speedup.
The framework is independent of a specific graphical interface and can be applied to other multimodal perception--reasoning--action loops.

\end{document}